\pdfoutput=1
\documentclass{article}

\usepackage[nonatbib, final]{neurips_2022}

\usepackage[utf8]{inputenc} 
\usepackage[T1]{fontenc}    
\usepackage{hyperref}       
\usepackage{url}            
\usepackage{booktabs}       
\usepackage{amsfonts}       
\usepackage{nicefrac}       
\usepackage{microtype}      
\usepackage{xcolor}         
\usepackage{multirow}
\usepackage{authblk}
\usepackage{graphicx}
\newcommand\cincludegraphics[2][]{\raisebox{-0.65\height}{\includegraphics[#1]{#2}}}

\title{Transformer Utilization in Medical Image Segmentation Networks}

\author[1]{\textbf{Saikat Roy}}
\author[1]{\textbf{Gregor Koehler}}
\author[1]{\textbf{Michael Baumgartner}}
\author[1]{\textbf{Constantin Ulrich}}
\author[1]{\textbf{Jens Petersen}}
\author[1,2]{\textbf{Fabian Isensee}}
\author[1,3]{\textbf{Klaus Maier-Hein}}

\affil[1]{Division of Medical Image Computing, German Cancer Research Center (DKFZ), Heidelberg, Germany \authorcr
  \{\tt saikat.roy, g.koehler, m.baumgartner, constantin.ulrich, f.isensee, j.petersen, k.maier-hein\}@dkfz-heidelberg.de}
\affil[2]{Helmholtz Imaging, German Cancer Research Center (DKFZ), Heidelberg, Germany
  }
\affil[3]{Pattern Analysis and Learning Group, Department of Radiation Oncology, Heidelberg University Hospital, Heidelberg, Germany 
  }

\begin{document}

\maketitle

\begin{abstract}
  Owing to success in the data-rich domain of natural images, Transformers have recently become popular in medical image segmentation. However, the pairing of Transformers with convolutional blocks in varying architectural permutations leaves their relative effectiveness to open interpretation. We introduce \textit{Transformer Ablations} that replace the Transformer blocks with plain linear operators to quantify this effectiveness. With experiments on 8 models on 2 medical image segmentation tasks, we explore -- 1) the replaceable nature of Transformer-learnt representations, 2) Transformer capacity alone cannot prevent representational replaceability and works in tandem with effective design, 3) The mere existence of explicit feature hierarchies in transformer blocks is more beneficial than accompanying self-attention modules, 4) Major spatial downsampling before Transformer modules should be used with caution.
\end{abstract}

\section{Introduction}
Vast gains have been achieved in recent years in natural language and computer vision domains with the availability of large datasets benefiting Transformer-based architectures in achieving state-of-the-art performance on various tasks \cite{khan2021Transformers,lin2021survey,han2022survey,chaudhari2021attentive}. Accordingly, Transformers have begun to be adapted at increasing pace in medical image segmentation with new architectures \cite{gao2021utnet, xu2021levit, hatamizadeh2022swin, zhou2021nnformer, petit2021u, hatamizadeh2021unetr} as well as training methods \cite{tang2021self} being routinely introduced. In this work, we attempt to quantify the relative Transformer utilization in 8 popular Transformer-based segmentation networks by means of \textit{Transformer Ablation}. This offers insight into a number of interlinked ideas in architecture design and the replaceable nature of Transformer-learnt representations, which we discuss alongside our results.

\section{Transformer Ablation}
The self-attention mechanism in Transformers can represented as  $X = s(QK^T) \cdot V$ where $Q,K,V \in \mathbb{R}^{N \times d}$, $s$ is a scaling function and $N$ and $d$ are length and dimensionality of the sequence vector. Local mechanisms such as those in Shifted-Window Transformers \cite{liu2021swin} restrict the self-attention computation to local windows, as opposed to the whole sequence as in the standard Vision Transformer (ViT) \cite{dosovitskiy2020image}. Transformer Ablation is defined as the removal of the Transformer block and replacement with a linear projection based tokenizer (in case of ViT) and linear projection with PatchMerging (in case of Swin-Transformers) to preserve downstream tensor compatibility. It is an extreme form of ablation designed to quantify the influence of Transformer-learnt representations in a network, \textit{by measuring the remaining network's ability to compensate} for performance in their absence. The ablation style of 5 out of 8 models under experimentation are illustrated in Table \ref{tab:architectures_abl}.

\begin{table*}[tb]

\centering  
\begin{tabular}{|cc|l|}
\hline
\textbf{Pre-ablation} & \textbf{Post-ablation} & \textbf{Networks} \\
\hline
\cincludegraphics[height=1.25cm, width=4cm]{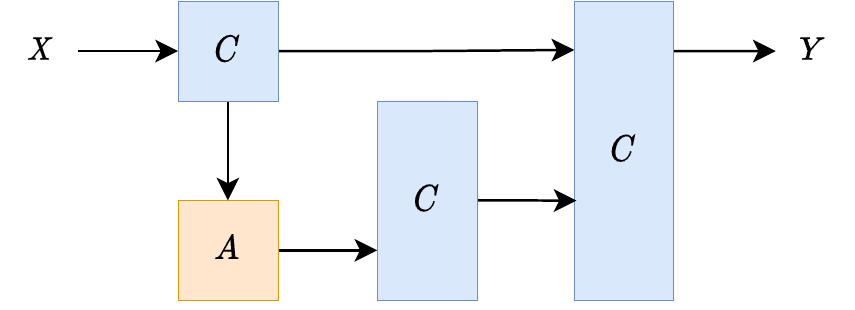} & \cincludegraphics[height=1.25cm,width=4cm]{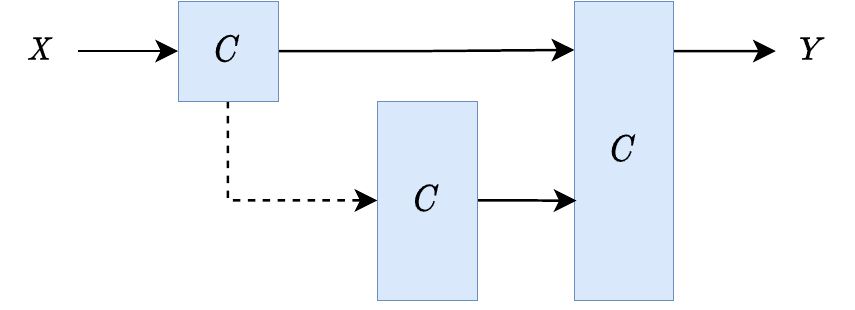} & \parbox[t]{2cm}{UNETR \\ SwinUNETR} \\ && \\
\cincludegraphics[height=1.25cm,width=4cm]{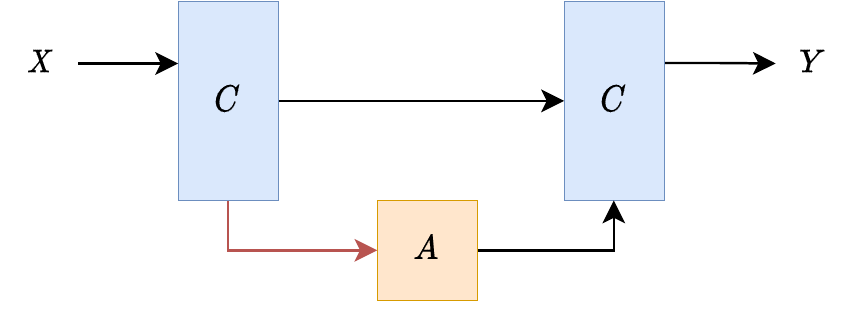} & \cincludegraphics[width=4cm]{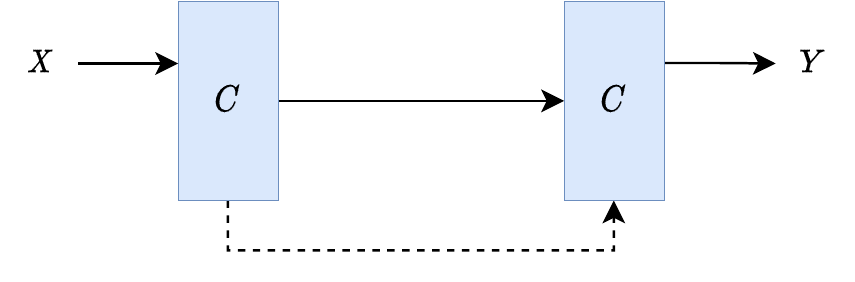} & \parbox[t]{2cm}{TransUNet \\ TransBTS} \\ 
\cincludegraphics[height=1.25cm,width=2.75cm]{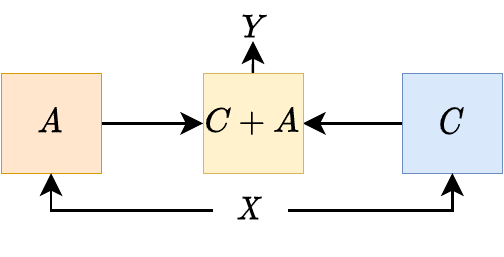} & \cincludegraphics[height=1.25cm,width=2.75cm]{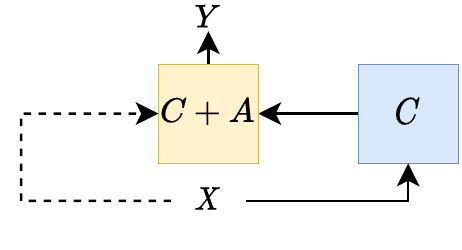} & TransFuse \\ 
\hline
\end{tabular}\
\caption{Pre and Post-\textit{Attention Ablation} forms of 5 out of 8 architectures in this work where $C$, $A$, $X$ and $Y$ denote Convolutions, Transformers, Input and Outputs respectively. Dotted paths denote \textit{compensatory} operations and \textcolor{red}{red} paths indicate a block having \textit{only} spatially downsampled inputs.}
\label{tab:architectures_abl}%
\end{table*}

\section{Experimental Design}
The Transformer blocks of 8 architectures (Table \ref{tab:mean_sd}) are ablated in this work and replaced with \textit{compensatory operations} for preserving tensor shape. The \textit{default} variant of each architecture is used, with exceptions being TransFuse (TransFuse-S) and nnFormer (Brain Tumor), while nnUNet \cite{isensee2021nnu} is used as a standard convolutional baseline. Results are obtained on 2 datasets -- a) Kidney Tumor Segmentation (KiTS) 2021 Dataset, b) Multi-Organ Abdominal CT (MultiACT) Dataset and compared against each other using 5-fold cross validation. Dice Similarity Coefficient (DSC) and Surface Dice Coefficient (SDC) with 1mm tolerance are used to compare network performance. KiTS2021 and MultiACT have dataset sizes of 300 and 90 volumes with 3 and 8 segmentable structures respectively. The networks use the nnUNet pipeline \cite{isensee2021nnu} for training with patch sizes of $128 \times 128 \times 128$ for 3D networks and $512 \times 512$ for 2D networks. The training pipeline is used unchanged in all aspects except for using AdamW \cite{loshchilov2017decoupled} as the optimizer instead of the standard SGD.  

\section{Results and Discussion}
The results obtained are illustrated in Table \ref{tab:mean_sd}, with relative change in trainable network parameters also highlighted. The following insights can be derived from the results:

\begin{table*}[t]
\centering
\begin{tabular}{l|c|c||c|c|c}
\hline
\multirow{2}{*}{\textbf{Model}} & \multicolumn{2}{c||}{\textbf{KiTS21}} & \multicolumn{2}{c|}{\textbf{MultiACT}} & \multirow{2}{*}{\textbf{Pars (in Mi)}} \\ \cline{2-5} 
 & \multicolumn{1}{c|}{\textbf{DSC}} & \textbf{SDC} & \multicolumn{1}{c|}{\textbf{DSC}} & \textbf{SDC} \\ \hline
UNETR \cite{hatamizadeh2021unetr} & 82.0 $\pm$ 1.0 & 68.9 $\pm$ 2.1 & 76.0 $\pm$ 2.1 & 64.6 $\pm$ 2.8 & 93.0 \\
\textit{Abl.} & 74.7 $\pm$ 3.1 & 59.0 $\pm$ 2.9 & 72.5 $\pm$ 1.8 & 59.5 $\pm$ 3.4 & 7.5 \\ \hline
\textit{Diff. $\mu$ or Ratio} & 7.3 & 9.9 & 3.5 & 5.1 & 0.08 \\ \hline \hline
TransBTS \cite{wang2021transbts} & 83.1 $\pm$ 2.4 & 72.6 $\pm$ 2.7 & 81.8 $\pm$ 2.0 & 72.7 $\pm$ 2.2  & 32.8 \\
\textit{Abl.} & 86.0 $\pm$ 1.0 & 76.0 $\pm$ 1.5 & 81.9 $\pm$ 1.9 & 73.1 $\pm$ 2.1 & 11.8 \\ \hline
\textit{Diff. $\mu$ or Ratio} & -2.9 & -3.4 & -0.1 & -0.4 & 0.36 \\ \hline \hline
SwinUNETR \cite{hatamizadeh2021unetr} & 84.3 $\pm$ 0.4 & 72.7 $\pm$ 2.3 & 79.8 $\pm$ 1.9 & 70.7 $\pm$ 2.2 & 15.7 \\
\textit{Abl.} & 84.3 $\pm$ 1.4 & 72.4 $\pm$ 2.9 & 78.7 $\pm$ 2.2 & 69.2 $\pm$ 2.2 & 14.3\\ \hline
\textit{Diff. $\mu$ or Ratio} & 0.0 & 0.3 & 1.1 & 1.5 & 0.91 \\ \hline \hline
CoTr \cite{xie2021cotr} & 88.6 $\pm$ 0.4 & 81.8 $\pm$ 0.5 & 84.2 $\pm$ 0.4 & 75.7 $\pm$ 0.4 & 41.9 \\
\textit{Abl.} & 88.0 $\pm$ 0.3 & 80.0 $\pm$ 0.3 & 83.1 $\pm$ 0.2 & 74.2 $\pm$ 0.2 & 32.6 \\ \hline
\textit{Diff. $\mu$ or Ratio} & 0.6 & 1.8 & 1.1 & 1.5 &  0.78 \\ \hline \hline
nnFormer \cite{zhou2021nnformer} & 79.8 $\pm$ 3.8 & 67.0 $\pm$ 3.4 & 80.3 $\pm$ 1.1 & 70.6 $\pm$ 2.4 & 37.6 \\
\textit{Abl.} & 84.2 $\pm$ 1.3 & 72.4 $\pm$ 1.9 & 81.3 $\pm$ 1.3 & 72.1 $\pm$ 2.3 & 6.1 \\ \hline
\textit{Diff. $\mu$ or Ratio} & -4.4 & -5.4 & -1.0 & -1.5 & 0.16 \\  \hline \hline
TransUNet \cite{chen2021transunet} & 81.5 $\pm$ 2.4 & 68.0 $\pm$ 1.7 & 82.8 $\pm$ 1.9 & 73.7 $\pm$ 2.3 & 105 \\
\textit{Abl.} & 81.6 $\pm$ 2.6 & 68.4 $\pm$ 1.9 & 83.0 $\pm$ 1.9 & 74.1 $\pm$ 2.4 & 20.9 \\ \hline
\textit{Diff. $\mu$ or Ratio} & -0.1 & -0.4 & -0.2 & -0.4 & 0.2 \\ \hline \hline
TransFuse \cite{zhang2021transfuse} & 81.7 $\pm$ 3.0 & 69.4 $\pm$ 1.6 & 81.7 $\pm$ 2.3 & 72.7 $\pm$ 3.1 & 26.0 \\
\textit{Abl.} & 82.5 $\pm$ 1.9 & 70.0 $\pm$ 0.6 & 83.6 $\pm$ 2.0 & 74.9 $\pm$ 3.1 & 11.8 \\ \hline
\textit{Diff. $\mu$ or Ratio}  & -0.8 & -0.6 & -1.9 & -2.2 & 0.45 \\ \hline \hline
UTNet \cite{gao2021utnet} & 81.4 $\pm$ 1.7 & 68.7 $\pm$ 0.9 & 82.9 $\pm$ 2.2 & 73.9 $\pm$ 2.9 & 10.0 \\
\textit{Abl.} & 81.4 $\pm$ 1.9 & 68.0 $\pm$ 1.4 & 82.4 $\pm$ 2.5 & 72.9 $\pm$ 3.0 & 8.1 \\ \hline
\textit{Diff. $\mu$ or Ratio} & 0.0 & 0.7 & 0.5 & 1.0 & 0.81 \\ \hline \hline
nnUNet \cite{isensee2021nnu} & 89.3 $\pm$ 0.7 & 80.8 $\pm$ 1.7 & 85.0 $\pm$ 1.1 & 78.2 $\pm$ 1.7 & - \\ \hline
\end{tabular}
\caption{Total learnable parameters as well as Mean \& Standard Deviation of DSC and SDC across 5 folds of all models on Standard (S) and \textit{Attention Ablation (Abl.)} modes on the KiTS21 and MultiACT datasets. Difference of means ($S-Abl.$) and ratio of params ($\frac{\#Abl.}{\#S}$) are also provided.}%
\label{tab:mean_sd}%
\end{table*}

\subsection{Replaceability of Representations}
When trained in conjunction with large convolutional or more efficient local attention (Swin) components, Transformers in medical image segmentation networks show a tendency to learn representations which can be replaced in their absence. This happens in almost all networks other than UNETR and to a smaller extent CoTr. This indicates that in a number of networks, the performance is driven by more data-efficient components whether they might be convolutions or Swin-blocks. This is not restricted to global ViTs -- in SwinUNETR, for example, Swin Transformer blocks are ablated from the network, resulting in minor changes to segmentation performance. However, maintained performance by the nnFormer abl., which is entire composed of Swin Transformer blocks show that the blocks themselves \textit{can learn usable representations}. We conclude that it is the architectural style and block combinations, not individual blocks themselves which result in the learning replaceable representations by modules.

\subsection{The role of Transformer vs Non-Transformer capacity}
Networks such as nnFormer retain segmentation performance on with only 16\% network parameters while CoTr suffers small degradations at even 78\% of original capacity. This indicates that the role of capacity is not straightforward and works in tandem with architecture design (eg., isolated ViTs in separate branch as in CoTr). Large capacity modules, in and around the Transformer (SwinUNETR where the Swin Transformer is only 9\% of the network), can often compensate for the representations learnt by them in their absence. Thus, network capacity must be treated as one aspect of a number of interlinked issues while designing Transformer-based medical image segmentation networks.

\subsection{The influence of Explicit Hierarchical Feature Learning}
Comparing the UNETR and SwinUNETR, it is seen that UNETR, while showing major performance degradation on losing its ViT (indicating higher Transformer utilization) is also less accurate overall than both SwinUNETR and its ablated form. The ablated SwinUNETR on the other hand, barely loses segmentation performance on losing its Swin Transformer and retaining PatchMerging operations for pooling. This highlights that the mere existence inductive bias in the form of explicit Hierarchical Feature Learning is highly beneficial to Transformer-based medical image segmentation networks. 

\subsection{Spatial Downsampling before Transformers in the bottleneck}
In the 2 bottleneck networks under investigation, TransBTS and TransUNet both retain performance during ablation. We theorize that besides the replaceability of representations, another factor at play might be the fact that both use significantly downsampled (8x) feature maps before their Transformer module which is designed to learn \textit{long-range dependencies}. We therefore advise caution against bottleneck Transformer designs with standard input sizes used in medical image segmentation.

\section{Conclusion}
The nature of Transformer-learnt representations in different medical image segmentation architectures are challenging to analyze. In this work, we explore them quantitatively by ablating the entire transformer and observing the ability of the remaining network to cope with the loss of the module. In doing so, we derive a number of insights into architecture design using Transformers in medical image segmentation, hoping to encourage better network designs and further work in this area.

\section{Potential Negative Societal Impact}
The analysis of Transformer-learnt representations holds no potential negative societal impact, as far as we are aware. Analysis of this nature, should enable researchers to design networks which maximize the potential of Transformers in their architectures.

\bibliographystyle{unsrt}
\bibliography{references}




\end{document}